\documentclass[10pt, a4paper]{article}
\usepackage{lrec}
\usepackage{multibib}
\newcites{languageresource}{Language Resources}
\usepackage{graphicx}
\usepackage{tabularx}
\usepackage{color}
\usepackage{soul}
\usepackage{epstopdf}
\usepackage[latin1]{inputenc}

\usepackage{hyperref}
\usepackage{xstring}
\usepackage{titlesec}
\titleformat{\section}{\normalfont\large\bf\center}{\thesection.}{1em}{}
\titleformat{\subsection}{\normalfont\SmallTitleFont\bf\raggedright}{\thesubsection.}{1em}{}
\titleformat{\subsubsection}{\normalfont\normalsize\bf\raggedright}{\thesubsubsection.}{1em}{}
\renewcommand\thesection{\arabic{section}}
\renewcommand\thesubsection{\thesection.\arabic{subsection}}
\renewcommand\thesubsubsection{\thesubsection.\arabic{subsubsection}}

\title{High Quality ELMo Embeddings for Seven Less-Resourced Languages}

\name{Matej Ul\v{c}ar, Marko Robnik-\v{S}ikonja}

\address{University of Ljubljana, Faculty of Computer and Information Science \\
         Ve\v{c}na pot 113, SI-1000 Ljubljana, Slovenia \\
         \{matej.ulcar, marko.robnik\}@fri.uni-lj.si\\}

\abstract{
Recent results show that deep neural networks using contextual embeddings significantly outperform non-contextual embeddings on a majority of text classification tasks. We offer precomputed embeddings from popular contextual ELMo model for seven languages: Croatian, Estonian, Finnish, Latvian, Lithuanian, Slovenian, and Swedish. We demonstrate that the quality of embeddings strongly depends on the size of the training set and show that existing publicly available ELMo embeddings for listed languages shall be improved. We train new ELMo embeddings on much larger training sets and show their advantage over baseline non-contextual fastText embeddings. In evaluation, we use two benchmarks, the analogy task and the NER task. \\ \newline 
}

\begin{document}

\maketitleabstract

\section{Introduction}
\textbf{Word embeddings} are representations of words in numerical form, as vectors of typically several hundred dimensions. The vectors are used as an input to machine learning models; for complex language processing tasks these are typically deep neural networks. The embedding vectors are obtained from specialized learning tasks, based on neural networks, e.g., word2vec \cite{mikolov2013exploiting}, GloVe \cite{pennington2014glove}, fastText \cite{Bojanowski2017}, ELMo \cite{Peters2018}, and BERT \cite{Devlin2019}.  For training, the embeddings algorithms use  large monolingual corpora that encode important information about word meaning as distances between vectors. In order to enable downstream machine learning on text understanding tasks, the embeddings  shall preserve semantic relations between words, and this is true even across languages. 

Probably the best known word embeddings are produced by the word2vec method \cite{mikolov2013distributed}. The problem with word2vec embeddings is their failure to express polysemous words. During training of an embedding, all senses of a given word (e.g., \emph{paper} as a material, as a newspaper, as a scientific work, and as an exam) contribute relevant information in proportion to their frequency in the training corpus. This causes the final vector to be placed somewhere in the weighted middle of all words' meanings.
Consequently, rare meanings of words are poorly expressed with word2vec and the resulting vectors do not offer good semantic representations. For example, none of the 50 closest vectors of the word \emph{paper} is related to science\footnote{This can be checked with a demo showing words corresponding to near vectors computed with word2vec from Google News corpus,  available at \url{http://bionlp-www.utu.fi/wv_demo/}.}. 

The idea of \textbf{contextual embeddings} is to generate a different vector for each context a word appears in and the context is typically defined sentence-wise. To a large extent, this solves the problems with word polysemy, i.e. the context of a sentence is typically enough to disambiguate different meanings of a word for humans and so it is for the learning algorithms. 
In this work, we describe high-quality models for contextual embeddings, called ELMo \cite{Peters2018}, precomputed for seven morphologically rich, less-resourced languages: Slovenian, Croatian, Finnish, Estonian, Latvian, Lithuanian, and Swedish. ELMo is one of the  most successful approaches to contextual word embeddings. At time of its creation, ELMo has been shown to outperform  previous word embeddings \cite{Peters2018} like word2vec and GloVe on many NLP tasks, e.g., question answering, named entity extraction, sentiment analysis, textual entailment, semantic role labeling, and coreference resolution.
While recently much more complex models such as BERT \cite{Devlin2019} have further improved the results, ELMo is still useful for several reasons: its neural network only contains three layers and the explicit embedding vectors are therefore much easier to extract, it is faster to train and adapt to specific tasks. 

This report is split into further five sections. In section \ref{sec:ELMo}, we describe the contextual embeddings ELMo. In Section \ref{sec:datasets}, we describe the datasets used,  and in Section \ref{sec:training} we describe preprocessing and training of the embeddings.   We describe the methodology for evaluation of created vectors and the obtained results in Section \ref{sec:evaluation}.
We present conclusion in Section \ref{sec:conclusions} where we also outline plans for further work.

\section{ELMo}
\label{sec:ELMo}
Standard word embeddings models or representations, such as word2vec \cite{mikolov2013exploiting}, GloVe \cite{pennington2014glove}, or fastText \cite{Bojanowski2017}, are fast to train and have been pre-trained for a number of different languages. They do not capture the context, though, so each word is always given the same vector, regardless of its context or meaning. This is especially problematic for polysemous words.
ELMo (Embeddings from Language Models) embedding \cite{Peters2018} is one of the state-of-the-art pretrained transfer learning models, that remedies the problem and introduces a contextual component. 

ELMo model`s architecture consists of three neural network layers. The output of the model after each layer gives one set of embeddings, altogether three sets. The first layer is a CNN layer, which operates on a character level. It is context independent, so each word always gets the same embedding, regardless of its context. It is followed by two biLM layers. A biLM layer consists of two concatenated LSTMs. In the first LSTM, we try to predict the following word, based on the given past words, where each word is represented by the embeddings from the CNN layer. In the second LSTM, we try to predict the preceding word, based on the given following words. The second LSTM is equivalent to the first LSTM, just reading the text in reverse.

In NLP tasks, any set of these embeddings may be used; however, a weighted average is usually employed. The weights of the average are learned during the training of the model for the specific task. Additionally, an entire ELMo model can be fine-tuned on a specific end task.

Although ELMo is trained on character level and is able to handle out-of-vocabulary words, a vocabulary file containing most common tokens is used for efficiency during training and embedding generation. The original ELMo model was trained on a one billion word large English corpus, with a given vocabulary file of about 800,000 words. Later, ELMo models for other languages were trained as well, but limited to larger languages with many resources, like German and Japanese. 

\subsection{ELMoForManyLangs}
\label{sec:efml}
Recently, ELMoForManyLangs \cite{che-EtAl:2018:K18-2} project released pre-trained ELMo models for a number of different languages \cite{fares-EtAl:2017:NoDaLiDa}. These models, however, were trained on significantly smaller datasets. They used 20-million-words data randomly sampled from the raw text released by the CoNLL 2017 Shared Task - Automatically Annotated Raw Texts and Word Embeddings \citelanguageresource{conll2017}, which is a combination of Wikipedia dump and common crawl. The quality of these models is questionable. For example, we compared the Latvian model by ELMoForManyLangs with a model we trained on a complete  Latvian corpus (wikidump + common crawl), which has about 280 million tokens. The difference of each model on the word analogy task is shown in Figure \ref{fig:latvian} in Section \ref{sec:evaluation}.  As the results of the ELMoForManyLangs embeddings  are significantly worse than using the full corpus, we can conclude that these embeddings are not of sufficient quality. For that reason, we computed ELMo embeddings for seven languages on much larger corpora. As this effort requires access to large amount of textual data and considerable computational resources, we made the precomputed models publicly available by depositing them to Clarin repository\footnote{\url{http://hdl.handle.net/11356/1277}}.   

\section{Training Data}
\label{sec:datasets}
We trained ELMo models for seven languages: Slovenian, Croatian, Finnish, Estonian, Latvian, Lithuanian and Swedish. To obtain high-quality embeddings, we used large monolingual corpora from various sources for each language. Some corpora are available online under permissive licences, others are available only for research purposes or have limited availability. The corpora used in training are a mix of news articles and general web crawl, which we preprocessed and deduplicated.  Below we shortly describe the used corpora in alphabetical order of the involved languages. Their names and sizes are summarized in Table \ref{tab:monolingual}.

\begin{table*}[htb]
\begin{center}

\begin{tabular}{llrr}
Language	& Corpora & Size & Vocabulary size \\ \hline 
Croatian	& hrWaC 2.1, Riznica, Styria articles & 1.95 & 1.4 \\
Estonian	& CoNLL 2017, Ekspress Meedia articles & 0.68 & 1.2 \\
Finnish  	& STT articles, CoNLL 2017, Ylilauta downloadable version & 0.92 & 1.3 \\
Latvian  	& CoNLL 2017 & 0.27 & 0.6 \\
Lithuanian	& Wikipedia 2018, DGT-UD, LtTenTen14 & 1.30 & 1.1 \\
Slovene 	& Gigafida 2.0 & 1.26 & 1.4 \\
Swedish  	& CoNLL 2017, STT articles & 1.68 & 1.2 \\
	\hline 
\end{tabular} 

\caption{The training corpora used. We report their size (in billions of tokens), and ELMo vocabulary size (in millions of tokens). }
\label{tab:monolingual}
\end{center}
\end{table*}

\textbf{Croatian} dataset includes hrWaC 2.1 corpus\footnote{\url{http://hdl.handle.net/11356/1064}} \cite{ljubesic-klubicka-2014-bs}, Riznica\footnote{\url{http://hdl.handle.net/11356/1180}} \cite{riznica}, and articles of Croatian branch of Styria media house, made available to us through partnership in a joint project\footnote{\url{http://embeddia.eu}}. hrWaC was built by crawling the .hr internet domain in 2011 and 2014. Riznica is composed of Croatian fiction and non-fiction prose, poetry, drama, textbooks, manuals, etc. The Styria dataset consists of 570,219 news articles published on the Croatian 24sata news portal and niche portals related to 24sata.

\textbf{Estonian} dataset contains texts from two sources, CoNLL 2017 Shared Task - Automatically Annotated Raw Texts and Word Embeddings\footnote{\url{http://hdl.handle.net/11234/1-1989}} \citelanguageresource{conll2017}, and news articles made available to us by Ekspress Meedia due to partnership in the project. Ekspress Meedia dataset is composed of  Estonian news articles between years 2009 and 2019. The CoNLL 2017 corpus is composed of Estonian Wikipedia and webcrawl.

\textbf{Finnish} dataset contains articles by Finnish news agency STT\footnote{\url{http://urn.fi/urn:nbn:fi:lb-2019041501}}, Finnish part of the CoNLL 2017 dataset, and Ylilauta downloadable version\footnote{\url{http://urn.fi/urn:nbn:fi:lb-2016101210}} \citelanguageresource{Ylilauta-dl_en}. STT news articles were published between years 1992 and 2018. Ylilauta is a Finnish online discussion board; the corpus contains parts of the discussions from 2012 to 2014.

\textbf{Latvian} dataset consists only of the Latvian portion of the ConLL 2017 corpus, which is composed of Latvian Wikipedia and general webcrawl of Latvian webpages.

\textbf{Lithuanian} dataset is composed of Lithuanian Wikipedia articles from 2018, Lithuanian part of the DGT-UD corpus\footnote{\url{http://hdl.handle.net/11356/1197}}, and LtTenTen\footnote{\url{https://www.sketchengine.eu/lttenten-lithuanian-corpus/}}. DGT-UD is a parallel corpus of 23 official languages of the EU, composed of JRC DGT translation memory of European law, automatically annotated with UD-Pipe 1.2. LtTenTen is Lithuanian web corpus made up of texts collected from the internet in April 2014 \cite{tenten}.

\textbf{Slovene} dataset is formed from the Gigafida 2.0 corpus \citelanguageresource{gigafida} of standard Slovene. It is a general language corpus composed of various sources, mostly newspapers, internet pages, and magazines, but also fiction and non-fiction prose, textbooks, etc.

\textbf{Swedish} dataset is composed of STT Swedish articles and Swedish part of CoNLL 2017. The Finnish news agency STT publishes some of its articles in Swedish language. They were made available to us through partnership in a joint project. The corpus contains those articles from 1992 to 2017.

\section{Preprocessing and Training}
\label{sec:training}
Prior to training the ELMo models, we sentence and word tokenized all the datasets. The text was formatted in such a way that each sentence was in its own line with tokens separated by white spaces. CoNLL 2017, DGT-UD and LtTenTen14 corpora were already pre-tokenized. We tokenized the others using the NLTK library\footnote{\url{https://www.nltk.org/}} and its tokenizers for each of the languages. There is no tokenizer for Croatian in NLTK library, so we used Slovene tokenizer instead.
After tokenization, we deduplicated the datasets for each language separately, using the Onion (ONe Instance ONly) tool\footnote{\url{http://corpus.tools/wiki/Onion}} for text deduplication. We applied the tool on paragraph level for corpora that did not have sentences shuffled and on sentence level for the rest. We considered 9-grams with duplicate content threshold of 0.9.

For each language we prepared a vocabulary file, containing roughly one million most common tokens, i.e. tokens that appear at least $n$ times in the corpus, where $n$ is between 15 and 25, depending on the dataset size. We included the punctuation marks among the tokens. We trained each ELMo model using the default values used to train the original English ELMo (large) model.

ELMo models were trained on machines with either two or three Nvidia GeForce GTX 1080 Ti GPUs. The training took roughly three weeks for each model. The exact time depended on the number of GPUs, size of the corpus, and other tasks running concurrently on the same machine.

\section{Evaluation}
\label{sec:evaluation}
We evaluated the produced ELMo models for all languages using two evaluation tasks: a word analogy task and named entity recognition (NER) task. Below, we first shortly describe each task, followed by the evaluation results.

\subsection{Word Analogy Task}
The word analogy  task was  popularized  by \newcite{mikolov2013distributed}.   The goal is to find a term $y$ for a given term $x$ so that the relationship between $x$ and $y$ best resembles the given relationship $a : b$.
There are two main groups of categories: 5 semantic, and 10 syntactic. To illustrate a semantic relationship in the category ''capitals and countries'', consider for example that the word pair $a : b$ is given as ``Finland : Helsinki''. The task is to find the term $y$ corresponding to the relationship ``Sweden : $y$'', with the expected answer being $y=$ Stockholm. In syntactic categories, the two words in a pair have a common stem (in some cases even same lemma), with all the pairs in a given category having the same morphological relationship. For example, in the category ``comparative adjective'', given the word pair ``long : longer'', we have an adjective in its base form and the same adjective in a comparative form. That task is to find the term $y$ corresponding to the relationship ``dark : $y$'', with the expected answer being $y=$ darker, that is a comparative form of the adjective dark. 

In the vector space, the analogy task is transformed into search for nearest neighbours using vector arithmetic, i.e. we compute the distance between vectors: d(vec(Finland), vec(Helsinki)) and search for the word $y$ which would give the closest result in distance  d(vec(Sweden), vec($y$)). 
In the analogy dataset the analogies are already pre-specified, so we are measuring how close are the given pairs. 
%In the evaluation below, we use analogy datasets for all tested languages based on the English dataset by \cite{mikolov2013efficient} \footnote{\url{http://download.tensorflow.org/data/questions-words.txt}}. Due to English-centered bias of this dataset, we used a modified dataset which was first written in Slovene language and then translated into other languages \cite{ulcar2019}\citelanguageresource{ulcar-analogy-clarin}. 
In the evaluation below we use analogy datasets by \newcite{ulcar2019}, which are based on the dataset by \newcite{mikolov2013efficient} and are available at Clarin repository \citelanguageresource{ulcar-analogy-clarin}.

As each instance of analogy contains only four words without any context, the contextual models (such as ELMo) do not have enough context to generate sensible embeddings. We tackled this issue with two different approaches. 

\subsubsection{Average over Word Embeddings}
In the first approach, we calculated ELMo embeddings for each token of a large corpus and then averaged the vectors of all the occurences of each word, effectively creating non-contextual word embeddings. For each language, we used language specific Wikipedia as the corpus. The positive side of this approach is that it accounts for many different occurences of each word in various contexts and thus provides sensible embeddings. The downsides are that by averaging we lose context information, and that the process is lengthy, taking several days per language. We performed this approach on three languages: Croatian, Slovenian and English. We used these non-contextual ELMo embeddings in the word analogy task in the same way as any other non-contextual embeddings. 

We used the nearest neighbor metric to find the closest candidate word. If we find the correct word among the $n$ closest words, we consider that entry as successfully identified. The proportion of correctly identified words forms a measure called accuracy@$n$, which we report as the result. 

In Table \ref{tab:analogyavg}, we show the results for different layers of ELMo models used as embeddings and their comparison with the baseline fastText embeddings. Among ELMo embeddings, the best result on syntactic categories are obtained by using the vectors after 2nd layer (LSTM1), while the best result on semantic categories are obtained using vectors after the 3rd layer of the neural model (LSTM2). Compared to fastText, the results vary from language to language. In English, fastText embeddings outperform ELMo in both semantic and syntactic categories. In Slovenian, ELMo embeddings outperform fastText embeddings, significantly so in syntactic categories. In Croatian, ELMo outperforms fastText on syntactic categories, but on semantic categories fastText is a bit better.

\begin{table}[!h]
\begin{center}
\begin{tabularx}{\columnwidth}{Xrrrr}
      &&&& \\
      Layer & category & Croatian & Slovenian & English \\
      \hline
      CNN      & semantic & 0.081 & 0.059 & 0.120  \\
               & syntactic & 0.475 & 0.470 & 0.454  \\
      LSTM1    & semantic & 0.219 & 0.305 & 0.376  \\
               & syntactic & 0.663 & 0.677 & 0.595  \\
      LSTM2    & semantic & 0.214 & 0.306 & 0.404  \\
               & syntactic & 0.604 & 0.608 & 0.545  \\
      fastText & semantic & 0.284 & 0.239 & 0.667  \\
               & syntactic & 0.486 & 0.437 & 0.626  \\
      \hline
\end{tabularx}
\caption{The embeddings quality measured on the word analogy task, using accuracy@1 score, where 200,000 most common words were considered. The embeddings for each word were obtained by averaging the embeddings of each occurence in the Wikipedia. Results are shown for each layer of ELMo model separately and are averaged over all semantic (sem) and all syntactic (syn) categories, so that each category has an equal weight (i.e. results are first averaged for each category, and then these results are averaged).}
\label{tab:analogyavg}
 \end{center}
\end{table}

\subsubsection{Analogy in a Simple Sentence}
In the second approach to analogy evaluation, we used some additional text to form simple sentences using the four analogy words, while taking care that their noun case stays the same. For example, for the words "Rome", "Italy", "Paris" and "France" (forming the analogy Rome is to Italy as Paris is to $x$, where the correct answer is $x=$France), we formed the sentence "If the word Rome corresponds to the word Italy, then the word Paris corresponds to the word France". We generated embeddings for those four words in the constructed sentence, substituted the last word with each word in our vocabulary and generated the embeddings again. As typical for non-contextual analogy task, we measure the cosine distance ($d$) between the last word ($w_4$) and the combination of the first three words ($w_2-w_1+w_3$). We use the CSLS metric \cite{Conneau2018} to find the closest candidate word ($w_4$). %If we find the correct word among the five closest words, we consider that entry as successfully identified. The proportion of correctly identified words forms a statistic called accuracy@5, which we report as the result.
%\hl{We already use accuracy@1 in the caption of Table 2. Maybe we shall move the above description to the first approach of testing analogies.}

\paragraph*{}
We first compare existing Latvian ELMo embeddings from ELMoForManyLangs project with our Latvian embeddings, followed by the detailed analysis of our ELMo embeddings.
We trained Latvian ELMo using only CoNLL 2017 corpora. Since this is the only language, where we trained the embedding model on exactly the same corpora as ELMoForManyLangs models, we chose it for comparison between our ELMo model with ELMoForManyLangs. In other languages, additional or other corpora were used, so a direct comparison would also reflect the quality of the corpora used for training. In Latvian, however, only the size of the training dataset is different. ELMoForManyLangs uses only 20 million tokens and we use the whole corpus of 270 million tokens. 

\begin{figure*}[!ht]
\centering
    \includegraphics[width=\textwidth]{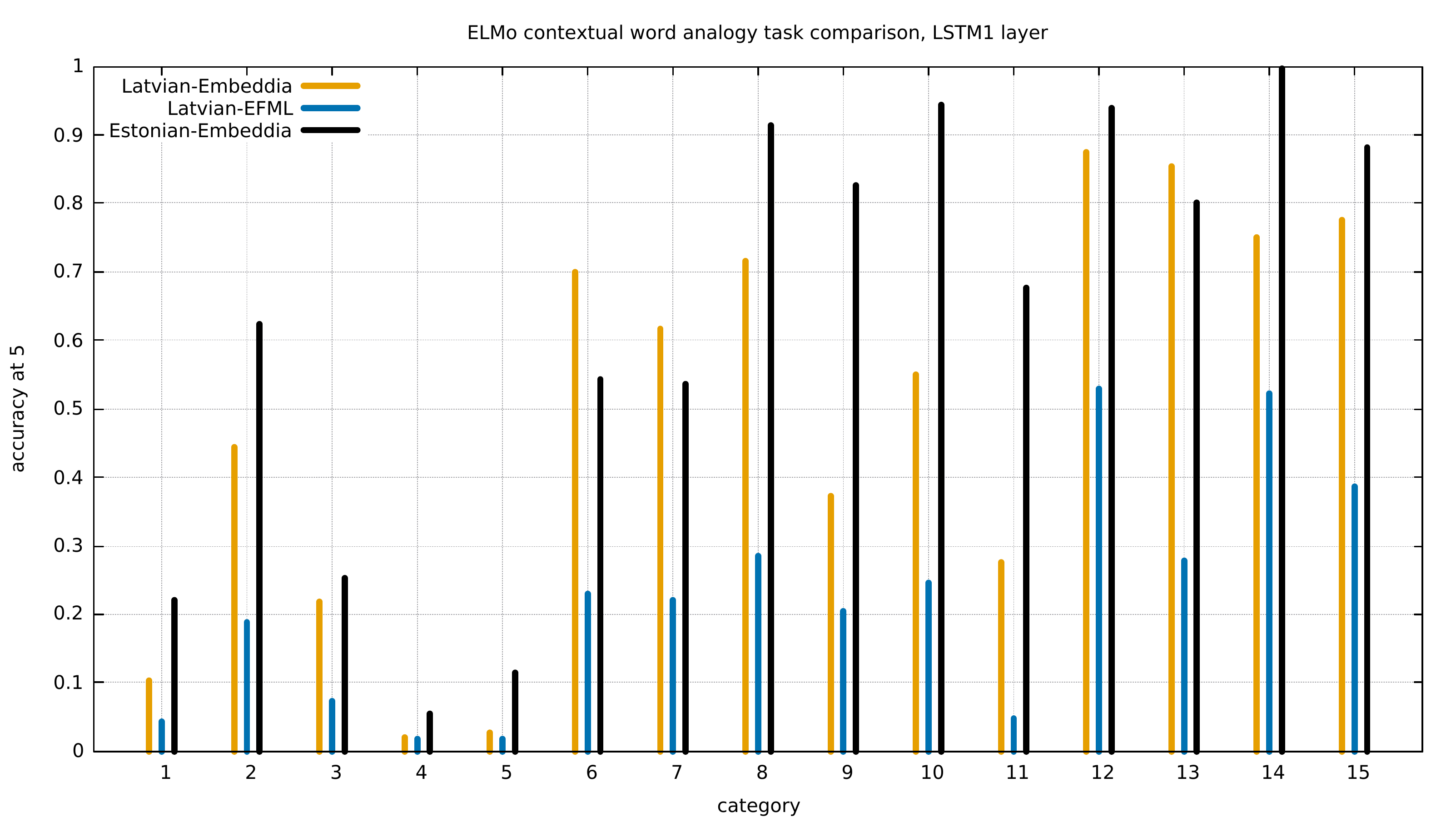}
    \caption{Comparison of Latvian ELMo model by ELMoForManyLangs (blue, Latvian-EFML), Latvian ELMo model trained by us (yellow, Latvian-Embeddia), and Estonian ELMo model trained by us (black, Estonian-Embeddia). The performance is measured as accuracy@5 on word analogy task, where categories 1 to 5 are semantic, and categories 6 to 15 are syntactic. The embeddings use weights of the first biLM layer LSTM1 (i.e. the second layer overall). %\hl{Could we shift x axis labels by 1, i.e. to odd numbers?}
    }
    \label{fig:latvian}
\end{figure*}

As Figure \ref{fig:latvian} shows, the Latvian ELMo model from ELMoForManyLangs project performs significantly worse than our ELMo Latvian model (named EMBEDDIA) on all categories of word analogy task. We also include the comparison with our Estonian ELMo embeddings in the same figure. This comparison shows that while differences between our Latvian and Estonian embeddings can be significant for certain categories, the accuracy score of ELMoForManyLangs is always worse than either of our models.
The comparison of Estonian and Latvian models leads us to believe that a few hundred million tokens forms a sufficiently large corpus to train ELMo models (at least for word analogy task), but 20-million token corpora used in ELMoForManyLangs are too small.

 The results for all languages and all ELMo layers, averaged over semantic and syntactic categories, are shown in Table \ref{tab:analogy}. The embeddings after the first LSTM layer (LSTM1) perform best in semantic categories. In syntactic categories, the non-contextual CNN layer performs the best. Syntactic categories are less context dependent and much more morphology and syntax based, so it is not surprising that the non-contextual layer performs well. The second LSTM layer embeddings perform the worst in syntactic categories, though they still outperform CNN layer embeddings in semantic categories. Latvian ELMo performs worse compared to other languages we trained, especially in semantic categories, presumably due to the smaller training data size. Surprisingly, the original English ELMo performs very poorly in syntactic categories and only outperforms Latvian in semantic categories. The low score can be partially explained by English model scoring $0.00$ in one syntactic category ``opposite adjective'', which we have not been able to explain. The English results strongly differ from the results of the first method (Table \ref{tab:analogyavg}). The simple sentence used might have caused more problems in English than in other languages, but additional evaluation in various contexts and other evaluation tasks would be needed to better explain these results. 

\begin{table}[!h]
\begin{center}
\begin{tabularx}{\columnwidth}{Xrrrrrr}
      &&&&&& \\
      Layer & \multicolumn{2}{c}{CNN} & \multicolumn{2}{c}{LSTM1} & \multicolumn{2}{c}{LSTM2} \\
      Category & sem & syn & sem & syn & sem & syn \\
      \hline
      hr & 0.13 & 0.79 & 0.24 & 0.75 & 0.20 & 0.54 \\
      et & 0.10 & 0.85 & 0.25 & 0.81 & 0.18 & 0.63 \\
      fi & 0.13 & 0.83 & 0.33 & 0.74 & 0.25 & 0.54 \\
      lv & 0.08 & 0.74 & 0.16 & 0.65 & 0.13 & 0.43 \\
      lt & 0.08 & 0.86 & 0.29 & 0.86 & 0.21 & 0.62 \\
      sl & 0.14 & 0.79 & 0.41 & 0.79 & 0.33 & 0.57 \\
      sv & 0.21 & 0.80 & 0.25 & 0.60 & 0.22 & 0.34 \\
      en & 0.18 & 0.22 & 0.21 & 0.22 & 0.21 & 0.21 \\
      \hline
\end{tabularx}
\caption{The embeddings quality measured on the word analogy task, using accuracy@5 score. Each language is represented with its 2-letter ISO code (first column). Results are shown for each layer separately and are averaged over all semantic (sem) and all syntactic (syn) categories, so that each category has an equal weight (i.e. results are first averaged for each category, and these results are then averaged).}
\label{tab:analogy}
 \end{center}
\end{table}

\subsection{Named Entity Recognition}
For evaluation of ELMo models on a relevant downstream task, we used named entity recognition (NER) task. NER is an information extraction task that seeks to locate and classify named entity (NE) mentions in unstructured text into pre-defined categories such as the person names, organizations, locations, medical codes, time expressions, quantities, monetary values, percentages, etc.
To allow comparison of results between languages, we used an adapted version of this task, which uses a reduced set of labels,  available in NER datasets for all processed languages.
The labels in the used NER datasets are simplified to a common label set of three labels (person - PER, location - LOC, organization - ORG). Each word in the NER dataset is labeled with one of the three mentioned labels or a label 'O' (Other, i.e. not a named entity) if it does not fit any of the other three labels. The number of words having each label is shown in Table \ref{tab:nertags}.

\begin{table}[!h]
\begin{center}

\begin{tabularx}{\columnwidth}{Xrrrrr}
      & & & & \\
      Language & PER & LOC & ORG & density & N \\ \hline
      %Croatian & 3931 & 656 & 1138 & 0.064 & & \\ %5725
      Croatian & 10241 & 7445 & 11216 & 0.057 & 506457 \\ %28902
      Estonian & 8490 & 6326 & 6149 & 0.096 & 217272 \\ %20965
      Finnish & 3402 & 2173 & 11258 & 0.087 & 193742 \\ %16833
      Latvian & 5615 & 2643 & 3341 & 0.085 & 137040 \\ %11599
      Lithuanian & 2101 & 2757 & 2126 & 0.076 & 91983 \\ %6984
      Slovenian & 4478 & 2460 & 2667 & 0.049 & 194667 \\ %9605
      Swedish & 3976 & 1797 & 1519 & 0.047 & 155332 \\ %7292
      English & 17050 & 12316 & 14613 & 0.146 & 301418 \\ %43979
%      Russian & 3293 & 2738 & 3635 & 0.107 \\
      \hline
\end{tabularx}
 
 \caption{The number of tokens labeled with each label (PER, LOC, ORG), the density of these labels (their sum divided by the number of all tokens) and the number of all tokens (N) for datasets in all languages.}
\label{tab:nertags}
\end{center}
\end{table}

To measure the performance of ELMo embeddings on the NER task we proceeded as follows. We split the NER datasets into training (90\% of sentences) and testing (10\% of sentences) set. We embedded text sentence by sentence, producing three vectors (one from each ELMo layer) for each token in a sentence. For prediction of NEs, we trained a neural network model, where we used three input layers (one embedding vector for each input). We then averaged the input layers, such that the model learned the averaging weights during the training. Next, we added two BiLSTM layers with 2048 LSTM cells each, followed by a time distributed softmax layer with 4 neurons.
%We embedded the text in the datasets sentence by sentence, producing three vectors (one from each ELMo layer) for each token in a sentence. %We calculated the average of the three vectors and used it as the input of our recognition model. The input layer was followed by a single LSTM layer with 128 LSTM cells and a dropout layer, randomly dropping 10\% of the neurons on both the output and the recurrent branch. The final layer of our model was a time distributed softmax layer with 4 neurons. 
\begin{figure*}[!ht]
\centering
    \includegraphics[width=\columnwidth]{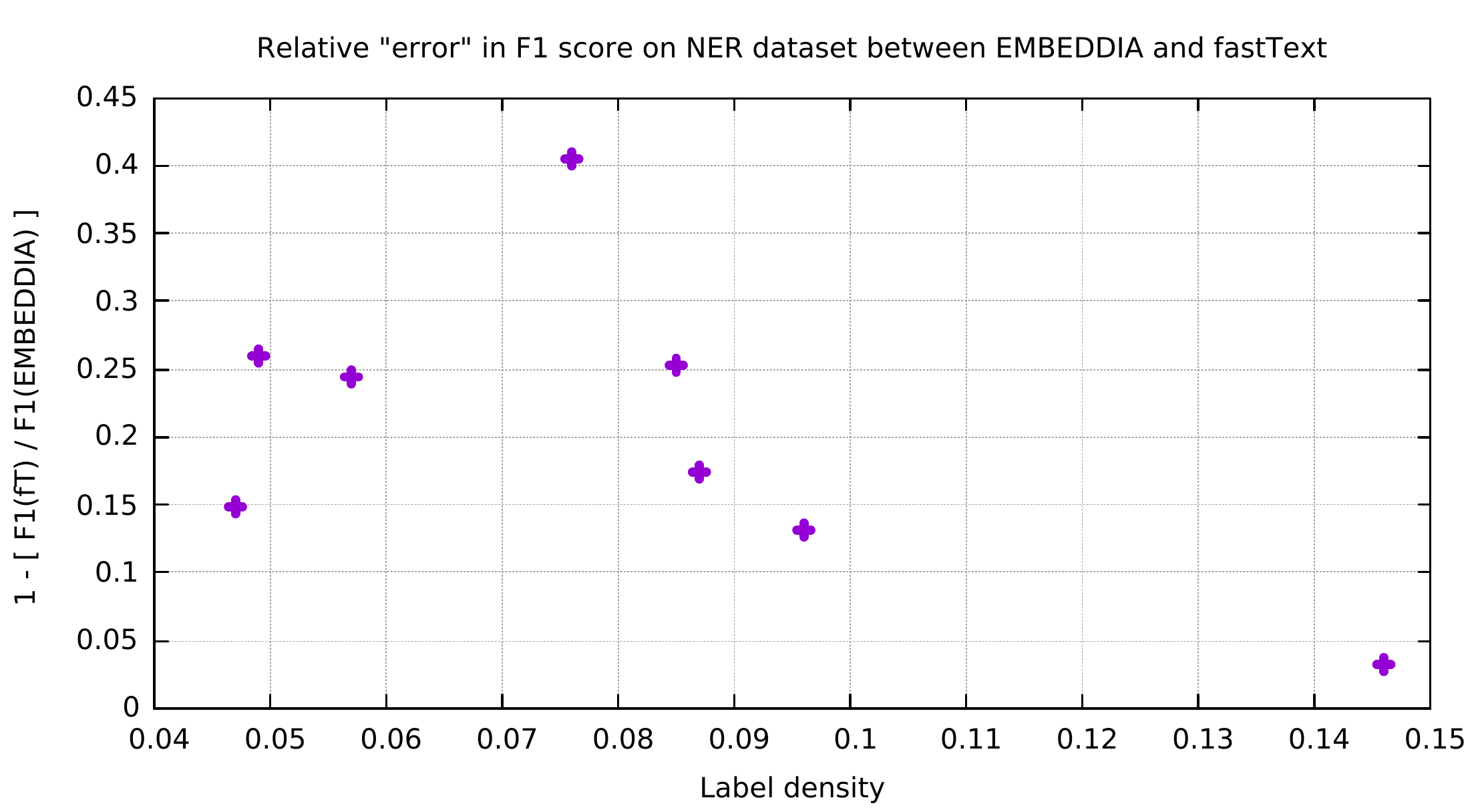}
    \includegraphics[width=\columnwidth]{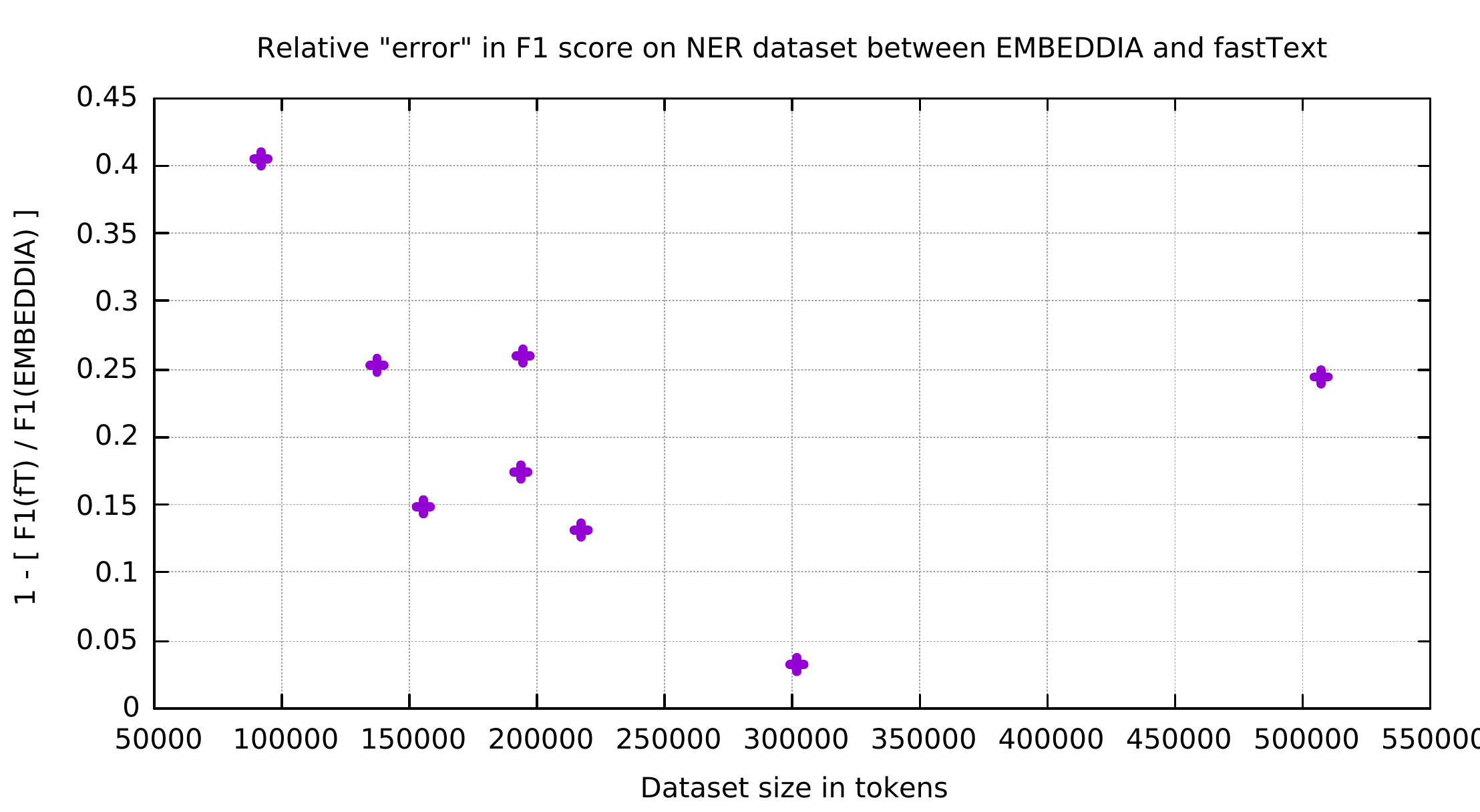}
    \caption{Comparison between fastText and EMBEDDIA ELMo embeddings on NER task. We show the relative difference (error) between the $F_1$ scores, in relation to the label density (left) and dataset size (right).
    }
    \label{fig:ftvsembeddia}
\end{figure*}

We used ADAM optimiser \cite{kingma2014adam} with the learning rate $10^{-4}$ and $10^{-5}$ learning rate decay. We used categorical cross-entropy as a loss function and trained each model for 10 epochs
(except Slovenian with EFML embeddings, where we trained for 5 epochs, since it gives a much better score ($0.82 F_1$ vs. $0.68 F_1$)).
We present the results using the Macro $F_1$ score, that is the average of $F_1$-scores for each of the three NE classes (the class Other is excluded) in Table \ref{tab:ner3}.

Since the differences between the  tested languages depend more on the properties of the NER datasets than on the quality of embeddings, we can not directly compare ELMo models. For this reason, we take the non-contextual fastText embeddings\footnote{\url{https://fasttext.cc/}} as a baseline and predict NEs using them. The architecture of the model using fastText embeddings is the same as the one using ELMo embeddings, except that we have one input layer, which receives 300 dimensional fastText embedding vectors. We also compared performance with ELMoForManyLangs (EFML) embeddings, using the same architecture as with our ELMo embeddings. In all cases (ELMo, EFML and fastText), we trained and evaluated prediction models five times and averaged the results due to randomness in initialization of neural network models. There is no Lithuanian EFML model, so we could not compare the two ELMo models on that language.

%The results are presented in Table \ref{tab:ner3}. We included the evaluation of the original ELMo English model in the same table. NER models have little difficulty distinguishing between types of named entities, but recognizing whether a word is a named entity or not is more difficult. For languages with the smallest NER datasets, Croatian and Lithuanian, ELMo embeddings show the largest  improvement over fastText embeddings. However, we can observe significant improvements with ELMo also on English and Finnish, which are among the largest datasets (English being by far the largest). Only on Slovenian dataset did ELMo perform slightly worse than fastText, on all other EMBEDDIA languages, the ELMo embeddings improve the results.
\begin{figure*}[!ht]
\centering
    \includegraphics[width=\columnwidth]{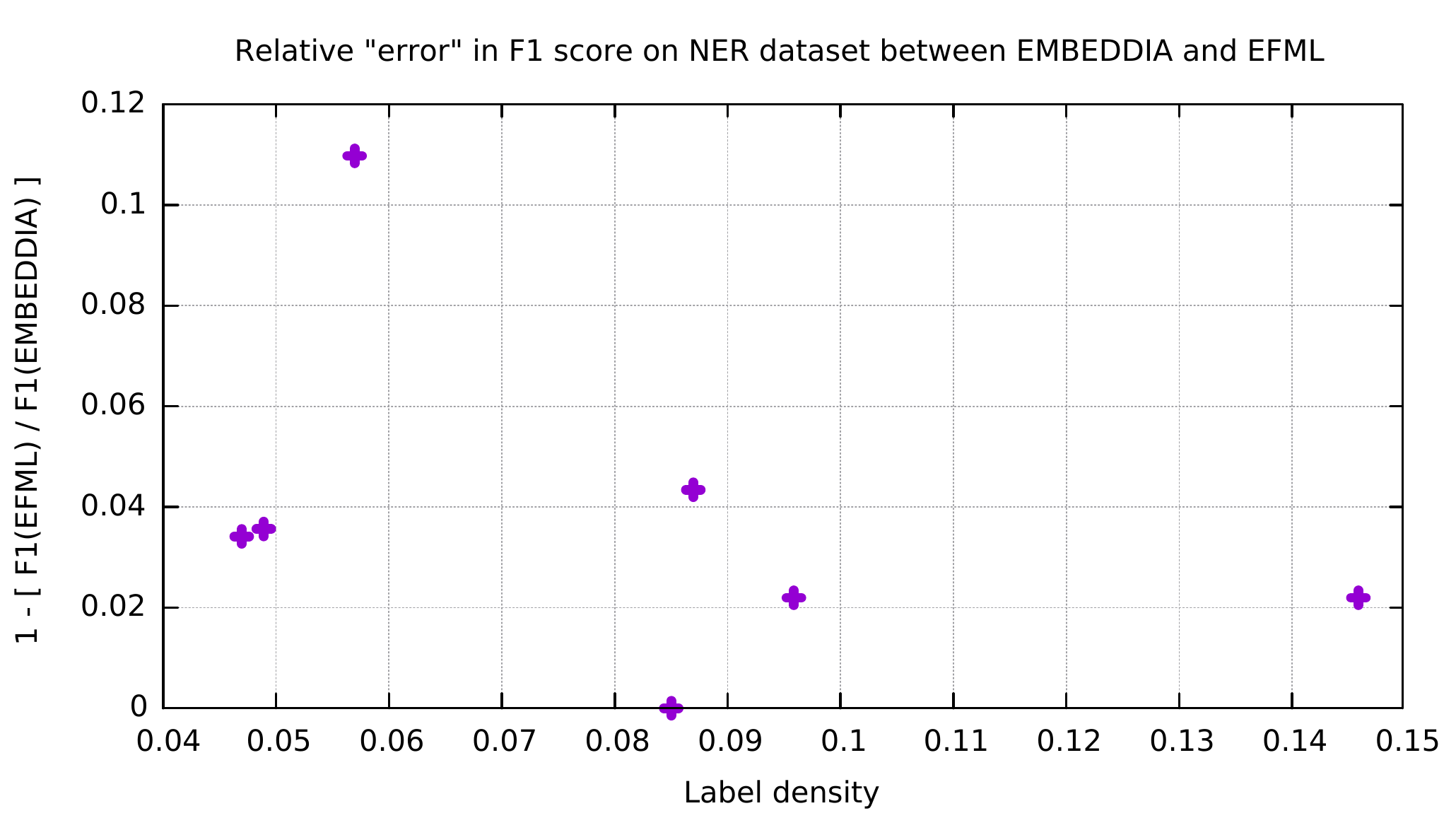}
    \includegraphics[width=\columnwidth]{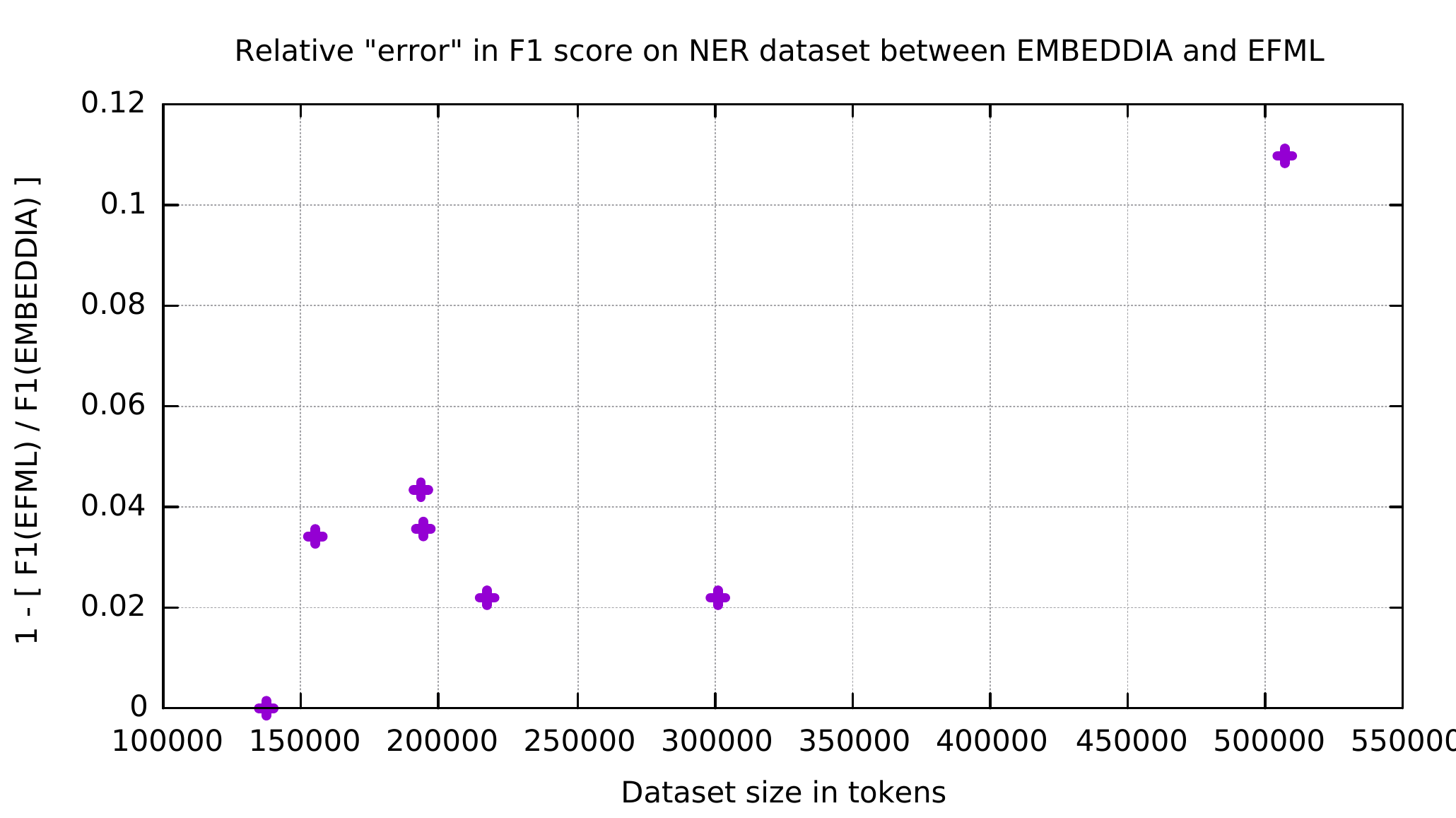}
    \caption{Comparison between EFML and EMBEDDIA ELMo embeddings on NER task. We show the relative difference (error) between the $F_1$ scores, in relation to the label density (left) and dataset size (right).
    }
    \label{fig:efmlvsembeddia}
\end{figure*}

Both ELMo embeddings (EFML and our EMBEDDIA) show significant improvement in performance on NER task over fastText embeddings on all languages, except English (Table \ref{tab:ner3}). In English, there is still improvement, but a smaller one, in part due to already high performance using fastText embeddings.

The difference between our ELMo embeddings and EFML embeddings is smaller on the NER task than on the word analogy task. On Latvian dataset, the performance is equal, while we have observed a significant difference on the word analogy task (Figure \ref{fig:latvian}). Our ELMo embedding models, however, show larger improvement over EFML on NER tasks in some other languages, like Croatian. 

We compared the difference in performance of EMBEDDIA ELMo embeddings and fastText embeddings as a function of dataset size and label density (Figure \ref{fig:ftvsembeddia}). Barring one outlier, there is a slight negative correlation with regard to the dataset size, but no correlation with label density. We compared the EFML and EMBEDDIA ELMo embeddings in the same manner (Figure \ref{fig:efmlvsembeddia}), with no apparent correlation.

%In comparison to fastText, the biggest improvement seem to be in morphologically rich languages with small datasets. Morphologically rich languages with larger datasets (Finnish and Estonian) show smaller, but still significant improvement. Similar observation can made for Swedish, which is not a morphologically rich language, but has a small dataset.
%\hl{Maybe the above claim is not evident as the sizes of the datasets are no clear from Table 4.}

\begin{table}[!h]
\begin{center}
\begin{tabularx}{\columnwidth}{Xrrr}
      & & & \\
      Language & fastText & EFML & EMBEDDIA \\ %& $\Delta (E-FT)$ \\
      \hline
      Croatian & 0.62 & 0.73 & 0.82  \\
      Estonian & 0.79 & 0.89 & 0.91  \\
      Finnish & 0.76 & 0.88 & 0.92  \\
      Latvian & 0.62 & 0.83 & 0.83  \\
      Lithuanian & 0.44 & N/A & 0.74  \\
      Slovenian & 0.63 & 0.82 & 0.85  \\
      Swedish & 0.75 & 0.85 & 0.88  \\
      English & 0.89 & 0.90 & 0.92  \\
      \hline
\end{tabularx}
 
 \caption{The results of NER evaluation task.  The scores are macro average $F_1$ scores of the three named entity classes, excluding score for class "Other". The columns show fastText, ELMoForManyLangs (EFML), and EMBEDDIA ELMo embeddings.}
\label{tab:ner3}
\end{center}
\end{table}

%\begin{table}[!h]
%\begin{center}

%\begin{tabularx}{\columnwidth}{Xrrr}
%      & OLD & ARCHITECTURE & RESULTS \\
%      Language & FastText & ELMo & $\Delta (E-FT)$ \\
%      \hline
%      Croatian & 0.17 & 0.53 & 0.36 \\
%      Estonian & 0.26 & 0.31 & 0.05 \\
%      Finnish & 0.71 & 0.84 & 0.13 \\
%      Latvian & 0.39 & 0.45 & 0.06 \\
%      Lithuanian & 0.43 & 0.65 & 0.22 \\
%      Slovenian & 0.68 & 0.67 & -0.01 \\
%      Swedish & 0.82 & 0.88 & 0.06 \\
%      English & 0.28 & 0.43 & 0.15 \\
%      \hline
%\end{tabularx}
 
% \caption{The results of NER evaluation task, averaged over 5 training and evaluation runs. The scores are average $F_1$ score of the three named entity classes. The columns show FastText, ELMo, and the difference between them ($\Delta (E-FT)$). }
%\label{tab:ner3}
%\end{center}
%\end{table}

\section{Conclusion}
\label{sec:conclusions}
We prepared high quality precomputed ELMo contextual embeddings for seven languages: Croatian, Estonian, Finnish, Latvian, Lithuanian, Slovenian, and Swedish. We present the necessary background on embeddings and contextual embeddings, the details of training the embedding models, and their evaluation. We show that the size of used training sets importantly affects the quality of produced embeddings, and therefore the existing publicly available ELMo embeddings for the processed languages can be improved for some downstream tasks. We trained new ELMo embeddings on larger training sets and analysed their properties on the analogy task and on the NER task. The results show that the newly produced contextual embeddings produce substantially better results compared to the non-contextual fastText baseline. In comparison with the existing ELMoForManyLangs embeddings, our new EMBEDDIA ELMo embeddings show a big improvement on the analogy task, and a significant improvement on the NER task. %especially when the task specific datasets are small.

For a more thorough analysis of our ELMo embeddings, more downstream tasks shall be considered. Unfortunately, no such task currently exist for the majority of the seven processed languages.

\paragraph*{}
As future work, we will use the produced contextual embeddings on the problems of news media industry. We plan to build and evaluate more complex models, such as BERT \cite{Devlin2019}. 
The pretrained EMBEDDIA ELMo models are publicly available on the CLARIN repository\footnote{\url{http://hdl.handle.net/11356/1277}}. 

\section{Acknowledgments} % po uradni predlogi je posebej \section
The work was partially supported by the Slovenian Research Agency (ARRS) through core research programme P6-0411 and research project J6-8256 (New grammar of contemporary standard Slovene: sources and methods). 
This paper is supported by European Union's Horizon 2020 research and  innovation programme under grant agreement No 825153, project EMBEDDIA (Cross-Lingual Embeddings for Less-Represented Languages in European News Media).
The results of this publication reflects only the authors' view and the EU Commission is not responsible for any  use that may be made of the information it contains.

% \nocite{*}
\section{Bibliographical References}
%\section*{References}
\label{main:ref}

\bibliographystyle{lrec}
\bibliography{lrec2020W-elmo}

%\section{Language Resource References}
\label{lr:ref}

\bibliographystylelanguageresource{lrec}
\bibliographylanguageresource{lrec2020W-elmo-lr}
%\bibliographystylelanguageresource{lrec}
%\bibliography{lrec2020W-elmo-lr}

\end{document}